\documentclass[letterpaper]{article} 
\usepackage[]{aaai25}  
\usepackage{times}  
\usepackage{helvet}  
\usepackage{courier}  
\usepackage[hyphens]{url}  
\usepackage{graphicx} 
\usepackage{natbib}  
\usepackage{caption} 
\usepackage{algorithm}
\usepackage{algorithmic}
\usepackage{newfloat}
\usepackage{listings}
\DeclareCaptionStyle{ruled}{labelfont=normalfont,labelsep=colon,strut=off} 
\floatstyle{ruled}
\newfloat{listing}{tb}{lst}{}
\floatname{listing}{Listing}
\nocopyright

\title{Benchmark on Peer Review Toxic Detection: \\A Challenging Task with a New Dataset}

\author {
    Man Luo\textsuperscript{\rm 1}\thanks{Equal contribution.},
    Bradley Peterson\textsuperscript{\rm 2}\footnotemark[1],
    Rafael Gan\textsuperscript{\rm 2},
    Hari Ramalingame\textsuperscript{\rm 2},
    Navya Gangrade\textsuperscript{\rm 2},
    Ariadne Dimarogona\textsuperscript{\rm 2},
    Imon Banerjee\textsuperscript{\rm 3},
    Phillip Howard\textsuperscript{\rm 1}
}
\affiliations {
    \textsuperscript{\rm 1} Intel Lab\\
    \textsuperscript{\rm 2} Arizona State University
    \textsuperscript{\rm 3} Mayo Clinic
}

\usepackage{bibentry}
\usepackage{booktabs} 
\usepackage{multirow} 
\usepackage{xcolor}

\begin{document}

\maketitle

\section*{Abstract}

Peer review is crucial for advancing and improving science through constructive criticism. However, toxic feedback can discourage authors and hinder scientific progress. This work explores an important but underexplored area: detecting toxicity in peer reviews.
We first define toxicity in peer reviews across four distinct categories and curate a dataset of peer reviews from the OpenReview platform, annotated by human experts according to these definitions. Leveraging this dataset, we benchmark a variety of models, including a dedicated toxicity detection model, a sentiment analysis model, several open-source large language models (LLMs), and two closed-source LLMs.
Our experiments explore the impact of different prompt granularities, from coarse to fine-grained instructions, on model performance. Notably, state-of-the-art LLMs like GPT-4 exhibit low alignment with human judgments under simple prompts but achieve improved alignment with detailed instructions. 
Moreover, the model's confidence score is a good indicator of better alignment with human judgments. For example, GPT-4 achieves a Cohen's Kappa score of 0.56 with human judgments, which increases to 0.63 when using only predictions with a confidence score higher than 95\%. 
Overall, our dataset and benchmarks underscore the need for continued research to enhance toxicity detection capabilities of LLMs. By addressing this issue, our work aims to contribute to a healthy and responsible environment for constructive academic discourse and scientific collaboration\footnote{Datset is available in this repository: \url{https://github.com/luomancs/toxic_peer_review_detection.git}}.

\section{Introduction}

Peer-review is essential to the editorial process and serves as the foundation of the publication system. 
It requires reviewers to have expertise in the research areas relevant to the submitted manuscripts. Reviewers are expected to provide constructive feedback in a polite and professional manner, facilitating a fair and productive assessment~\cite{mavrogenis2019evaluate,mavrogenis2020good}. 
A good peer-review can support and improve scientific communication and the dissemination of research.
However, the peer review process can inadvertently lead to the occurrence of toxic reviews if not handled with ethical consideration. Such reviews, characterized by harsh and unconstructive criticism, rude comments, and unprofessionalism, can severely impact authors, causing feelings of discouragement, anger, or even depression~\cite{baron1988negative,mavrogenis2020good}. This, in turn, may hinder scientific progress~\citep{ffdd70d0c1784f26aad78a19af959025,gerwing2020quantifying}. 
The prevalence of this issue is underscored by a survey in which 58\% of 1,106 participants reported having received an unprofessional review~\citep{silbiger2019unprofessional}.

Despite the importance of addressing toxicity in peer reviews, this area has received relatively little attention in current literature. Most efforts in the domain of toxicity detection are focusing on social media comments~\cite{cheng2022bias,hosseinmardi2015detection,ghosh2021detoxy,bensalem2024toxic}. As a result, there is a urgent need for systematic approaches to detect and mitigate toxic content in peer reviews.
Our research aims to fill this gap by introducing the first dataset to detect toxic sentences in peer reviews. 

To create a high-quality dataset, we begin by collecting a suit of reviews from the Open Review platform. 
One challenge is how to annotate the collected review since the definition of toxic review are not clearly documented. 
Therefore, we first define four characteristics of sentences that have high risk of being toxic by surveying a list of literature~\citep{silbiger2019unprofessional,mavrogenis2019evaluate,rogers2020can}, scientific interview~\footnote{\url{https://www.nature.com/nature-index/news/linda-beaumont-research-journals-should-take-action-against-toxic-peer-reviews}} and analyze the conference review gudeline such as ACL Rolling Review Guideline~\footnote{\url{https://aclrollingreview.org/reviewertutorial#6-check-for-lazy-thinking}},  which focus on the instruction of writing a good review and the clues of bad reviews. 
Then, five undergraduate annotators firstly identify potential toxic sentences from randomly selected reviews. 
Such sentences are further judged by senior researchers. 
To ensure the reliability of the dataset, the annotation procedure is conducted into two phases, and each phase consist of an independent judgement and a discussion among the annotators. Eventually only sentences with consenting from all annotators are included in our testing set. 
To this end, we have collect a high quality dataset with 313 sentences for testing. 

Using our dataset, we benchmarked multiple models, including toxicity detection model\footnote{We use the perspective API.}, sentiment analysis models~\cite{sanh2019distilbert}, and large language models (LLMs)~\cite{jiang2023mistral,touvron2023llama}, to perform toxic review detection task. We evaluated the toxic detection performance using precision, recall, F1 score, and Cohen's Kappa score, comparing these results with human judgments.
Our findings reveal that existing toxicity detection models struggle to accurately identify toxic content within the peer-review context. Conversely, a sentiment analysis model achieved higher alignment with human judgments compared to most open-source LLMs. Among the models tested, the closed-source model, particularly GPT-4, demonstrated the best performance. We further explored the impact of prompts on GPT-4's performance, discovering that detailed instructions significantly improve the model's performance and alignment with human judgments.
Additionally, we prompted the model to generate a confidence score with its predictions. Our results show that this confidence score is a reliable indicator of better alignment with human judgments, with a 7\% improvement in alignment when the confidence threshold is set at 95\%.
Beyond detection, we also explored the potential of LLMs to revise toxic sentences. We presented both the original and revised sentences to human evaluators, who reported that 80\% of the revisions were preferable, indicating a potential possibility of using LLMs to detoxify peer reviews.

Recognizing the absence of a formal definition for toxic reviews and the need for standardized preventative measures, our research aims to highlight the importance of this issue and contribute to the discourse by establishing key criteria for preventing toxic reviews. The four characteristics we have identified can serve as the foundation for developing guidelines in reviewer training programs, helping to preemptively mitigate toxic reviews or significantly reduce their occurrence. Furthermore, our experiments with LLM suggests their applications in post-review detection: they can identify potentially toxic language to a moderate degree. Ultimately, our goal is to foster a healthy and supportive research environment, enhancing both the mental well-being of individuals and the overall advancement of the research community.
To summarize our contribution in this paper: 
\begin{itemize}
\item We define four characteristic of toxic sentences in peer-review context as no previous work have provided comprehensive definition. We acknowledge that our definition might not be the most precise and can be improved further, but if a sentence falls into any of the categories, have high potential risk of being toxic. 
\item We develop the first benchmark dataset for detecting toxic sentences in peer-review context. This dataset follows rigorous human annotation to ensure its reliability.
\item We benchmark the performance of multiple models including existing toxicity classification model, sentiment analysis models, and advanced LLMs on our dataset and the results suggests that the open-source models are all struggling with our task. On the other hand, with proper instruction, the open-source models achieve acceptable agreement with human (e.g. 0.54 Cohen's Kappa Score). 
\end{itemize}

\section{Related Work}

Toxicity detection has been mostly studied  in social media and speech content, often interchangeably referred to as hate speech detection. 
Toxicity encompasses a range of harmful behaviors including being ``hateful'', ``abusive'', and ``cyberbullying''~\cite{cheng2022bias}. 
Two notable datasets include the Jigsaw and Instagram datasets. 
Jigsaw dataset\footnote{\url{https://www.kaggle.com/c/jigsaw-unintended-bias-in-toxicity-classification}} comprises comments extracted from the Civil Comments platform, annotated for toxicity and identity-related biases. This dataset includes various subsets, addressing issues such as unintentional bias and multilingual annotations, making it a valuable resource for studying toxic behavior across different languages and contexts. 
The Instagram dataset~\cite{hosseinmardi2015detection}, on the other hand, is derived from one of the most popular social networking sites where users frequently report experiences of cyberbullying. Each sample in this dataset represents a social media session composed of a sequence of comments in temporal order, providing insights into the dynamics of toxic interactions over time. In addition to these, the DeToxy dataset~\cite{ghosh2021detoxy} presents a large-scale multimodal collection specifically for toxicity classification in spoken utterances. This dataset expands the scope of toxicity detection beyond text-based content to include audio, highlighting the multimodal nature of online abuse. \citet{bensalem2024toxic} integrates 54 Arabic datasets into a unified format, facilitating easier access and comprehensive analysis for toxic comment classification. 
Additionally, toxicity detection in the gaming sector~\footnote{\url{https://www.databricks.com/blog/2021/06/16/solution-accelerator-toxicity-detection-in-gaming.html}} has garnered attention due to the impact of toxic behavior on player experiences and community health. Despite these advances, the application of toxicity  in the context of scientific paper reviews remains underexplored, indicating a clear gap that our research aims to address.



In contrast to the extensive work done on toxicity detection, toxicity revision, especially in the realm of scientific writing, is somewhat less-studied. Automated revision involving technical concepts is a particularly complex task, beyond the already-challenging task of re-writing for clarity or grammar alone. Jiang, Xu, and Stevens (2022) introduced the arXivEdits dataset, which documents human revisions in scientific papers, categorizing changes like content alterations, grammar corrections, and stylistic improvements. While this work sheds light on general revision practices, it does not specifically address toxic language in peer reviews. Our work aims to fill some of these gaps by investigating how LLMs can both detect and revise toxic language in scientific peer reviews.


\section{Paper Review Toxicity Dataset}

\subsection{Paper Review Collection }
To collect paper reviews, we utilized the OpenReview API, starting with a comprehensive list of all venues between 2018 and 2023, including conferences and workshops. 
Consequently, we successfully collected 50,108, spanning 47 unique conferences and workshops 
Among the total collection, we randomly sample 1,495 for the annotations, and 
on average, each review consisted of 1,553 characters. However, due to the heavy human annotations work, we are not using all the processed review. The distribution of the final testing set will be discussed in \S\ref{sec:human_annotation}.

\subsection{Toxicity Review Guideline}
\label{sec:guideline}

Many journals have guidelines and codes of conduct for peer-reviewers,  but the unacceptable behaviour are often difficult to be found on journal websites~\footnote{\url{https://www.nature.com/nature-index/news/linda-beaumont-research-journals-should-take-action-against-toxic-peer-reviews}}. 
ACL Rolling Review is an exception, where a list of lazy review behavior have been mentioned~\footnote{\url{https://aclrollingreview.org/reviewertutorial#6-check-for-lazy-thinking}}. Nevertheless, ``toxic'' behavior have not been clear defined and nevertheless become a norm of peer-review. 
Therefore, from multiple different sources (professional interview, scientific paper, human studies), we have summarized the unacceptable behaviors as follows to define a sentence that can be toxic.  

\paragraph{Emotive Comments.}  
Using emotive or sarcastic language is often the hardest for authors to cope with~\footnote{\url{https://www.nature.com/articles/d41586-020-03394-y}}. 
Such language include the use of the following words or phrases: speaker-oriented adverbs such as surprisingly, obviously and disappointingly, subjective adjectives (e.g. careless), depreciatory modifiers (e.g. even, just), unnecessary expression including narrativizing: (e.g. , ``At this point, I almost stopped reading''), rhetorical questioning (for example, ``Did the authors even read the submission guidelines?''), universalizing (for example, ``As anyone/everyone/any expert knows''), speculating (for example, ``I bet the outlier observations were omitted''), expressive punctuation, such as exclamation marks or scare quotes (for example, ``This is not correct!'').
Similarly, in a professional interview, an experienced journal editor has mentioned that a comment with personal emotion can be toxic, such as using the world ``silly'', or phrase `` they have no idea what they’re doing''. 

\paragraph{Lack of Constructive Feedback.} 
Refers to a situation where feedback provided to someone lacks substance, helpfulness, or guidance for improvement.
Instead of offering specific suggestions, actionable advice, or positive reinforcement, this type of feedback may be vague, overly critical, or unproductive. 
Example1: \textit{``Comparison in experiments looks meaningless.''}
Example2: \textit{``The analysis is shallow.''}
In addition, if the criticism is lack of reference or citations, such sentences are falling into this categories. Example3: \textit{``The paper is not well written and organized.''} 
The reviewer should refer to what problems happens, an alternative sentence with the same criticism can be \textit{``The paper is not well written and organized, examples will be given below.''}

\paragraph{Personal Attack.} 
are directed at the author(s) rather than the nature or quality of the work~\cite{silbiger2019unprofessional}.  Example1: \textit{``The author’s previous work has been unreliable, and this paper is no different. They should find a different career.''} Example2: \textit{``Perhaps the authors are not used to submitting to this conferences as the paper lacks some essential components to it.''}

\paragraph{Excessive Negativity.} A sentence that overly emphasizes flaws of a paper without acknowledging any merits, which can contributes to a toxic review environment. This usually happens in the summary or conclusion of a review.
Example1: \textit{"The paper is not properly written nor well organized; is hard to read with vague contributions and vague positioning with respect to the state of the art."} 
Example2: \textit{``This paper has no significance."}
Example3: \textit{``This paper cannot be revised so as to be suitable for our journal.''}

\subsection{Human Annotation}
\label{sec:human_annotation}
We designed a rigorous annotation process to ensure the reliability of our dataset. Initially, five undergraduate students majoring in computer science identified toxic sentences. These annotations were then verified by three senior researchers. To further enhance agreement among the senior researchers, we conducted a two-phase annotation process: first with 50 sentences, followed by a larger set of 300 sentences.
In each phase, the annotation process was consistent: each annotator independently rated the sentences, after which they discussed any sentences where there was disagreement and could adjust their ratings if necessary. Conducting a smaller set of annotations served two purposes: first, it helped the annotators reach substantial agreement before moving on to the more time-consuming larger set; second, based on Cohen's Kappa score, the two annotators with the highest agreement were selected to annotate the larger set of 300 sentences.
In the initial round of 50 sentences, the Cohen's Kappa scores among the three annotators were 0.74, 0.52, and 0.34, with a Fleiss' Kappa score of 0.55. After discussing the sentences with disagreements, the Fleiss' Kappa score increased to 0.63, and the highest Cohen's Kappa score between two annotators rose to 0.83. The two annotators with the highest Kappa scores were then assigned an additional 300 sentences. In this larger set, the initial Cohen's Kappa score was 0.60, which significantly increased to 0.92 after discussion.
In the experimental section, we will analyze the sentences where the two reviewers failed to reach a final agreement. Ultimately, we combined 36 sentences where all three annotators agreed in the first round and 277 sentences where the two selected annotators agreed in the second round, resulting in a testing set comprising 182 non-toxic and 131 toxic sentences (Table \ref{tab:statistic}). 

\begin{table}[t]
\centering
 \resizebox{0.60\linewidth}{!}{
\begin{tabular}{c|c|c}
    \toprule
     Total & Toxic & Non-Toxic \\
     \midrule
     313 & 131 & 182 \\
    \bottomrule
    \end{tabular}
    }
    \caption{Toxic Review Detection Dataset. The statisic is in sentence level.}
\label{tab:statistic}
\end{table}

\begin{figure}[h]
    \centering
    \includegraphics[width=0.8\linewidth]{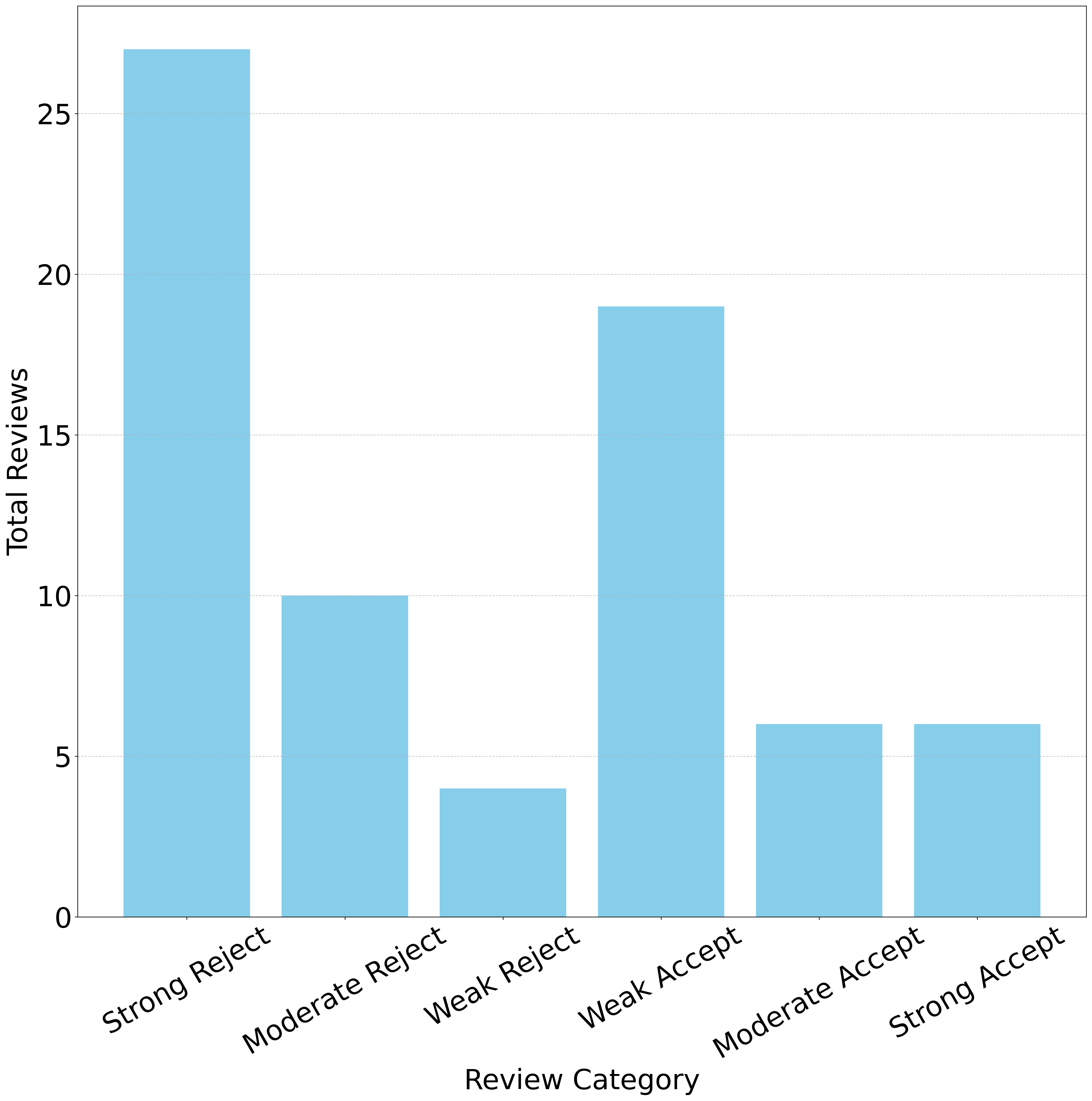}
    \caption{Distribution of Six Decision Categories of the Reviews of the \underline{Entire Testing Data}.}
    \label{fig:review_categories_1}
\end{figure}

\begin{figure}[h]
    \centering
    \includegraphics[width=0.8\linewidth]{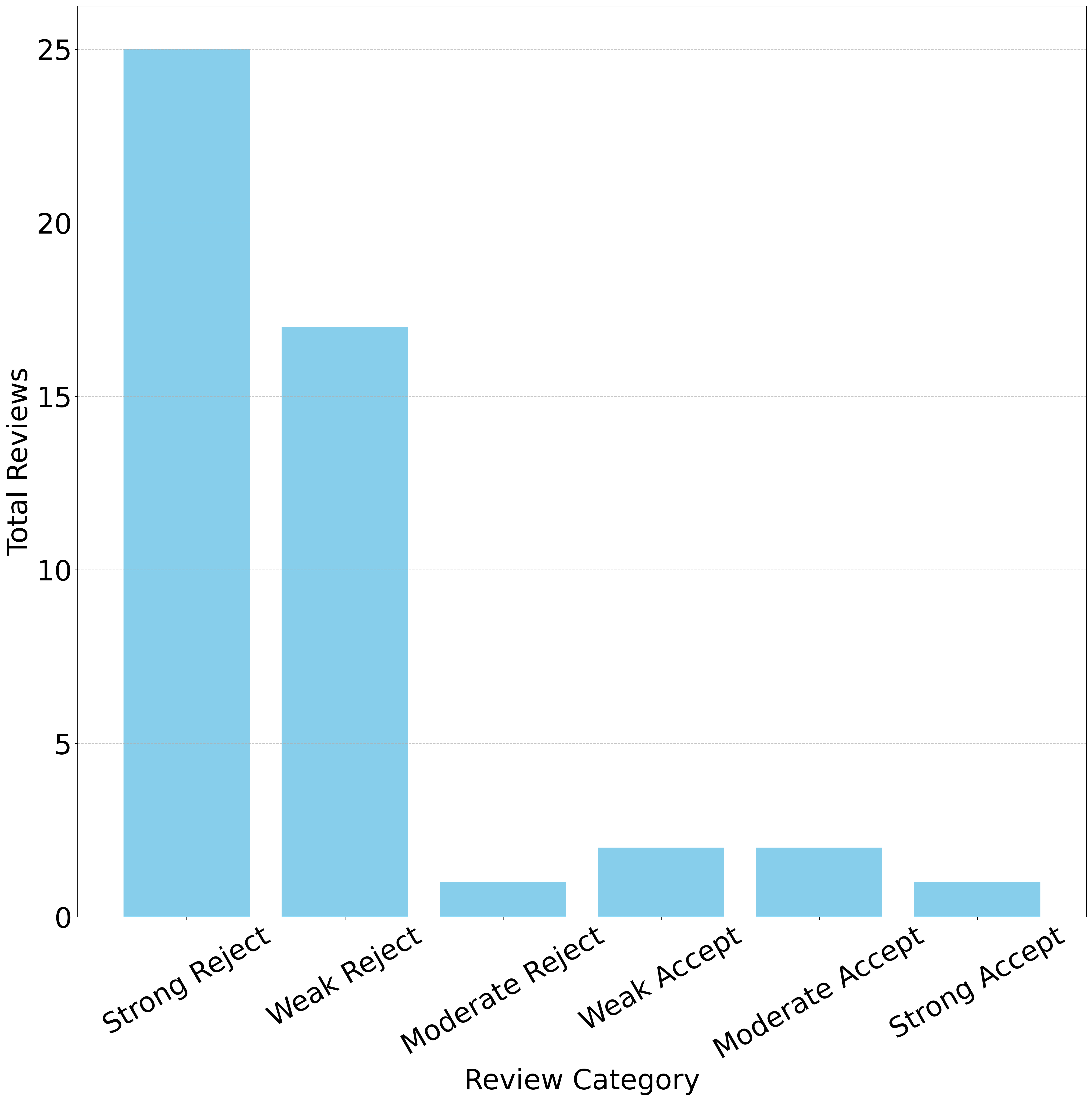}
    \caption{Distribution of Six Categories of the Reviews of the \underline{Toxic Sentences}.}
    \label{fig:review_categories_2}
\end{figure}

\paragraph{Review Distribution.}


The reviews of the sentences included in our final testing set are categorized into five groups: Strong Reject, Moderate Reject, Weak Reject, Weak Accept, Moderate Accept, and Strong Accept. The distribution of these categories within our final testing set is illustrated in Figure~\ref{fig:review_categories_1}. Additionally, Figure~\ref{fig:review_categories_2} shows the distribution of decision categories specifically for toxic sentences, revealing a shift toward more reject decisions. Even though sentences are analyzed individually, this trend suggests that reviews with overall reject decision receive more toxic comment which is well align with our expectation but could be extremely discouraging for the authors.

\begin{figure}[h]
    \centering
    \includegraphics[width=0.8\linewidth]{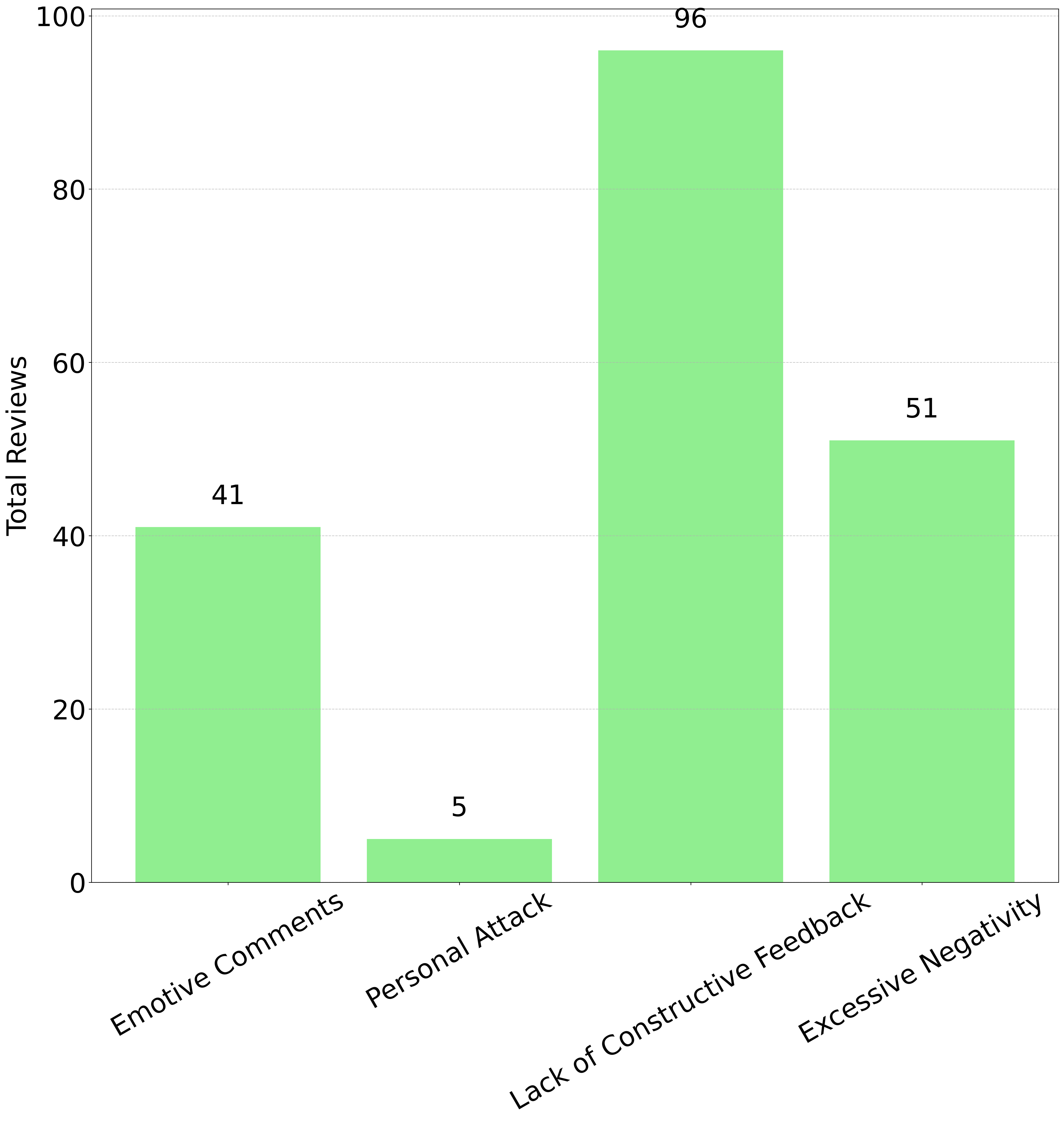}
    \caption{The Distribution of  Toxic-Subcategories.}
    \label{fig:toxic_categories}
\end{figure}

\paragraph{Subcategories Distribution.}
We present the distribution of  toxic-subcategories in Figure~\ref{fig:toxic_categories}.
Note that a toxic sentence could belong to multiple categories. 
The most common category is ``lack of constructive feedback'', follows by ``Excessive Negativity'', ``Emotive Comments'', and ``Personal Attack''. 

\paragraph{Discussion Among the Annotators.} 
The main discussion focuses on two key questions: 1) how to differentiate between assertive tones and emotive language, and 2) what constitutes a ``lack of constructive feedback.''

For the first question, some ambiguous sentences include ``Would a superior method here not just be to take this brain information directly?'' and ``The authors lack any clear discussion on how their work is directly relevant to RRL.'' In the first example, one annotator identified "just" as a depreciatory modifier, which falls under the category of \underline{Emotive Comments} (See \S\ref{sec:guideline}). Removing this word does not change the criticism. In the second example, the word ``any'' was discussed, with one annotator noting that removing it would weaken the criticism, categorizing it as an assertive tone. After discussion, the annotators agreed that if a word does not belong to any of the categories outlined in \underline{Emotive Comments} and its removal weakens the criticism, it should be classified as assertive rather than toxic.

For the second question, some ambiguous sentences were ``The proposed copy mechanism is not clear,'' ``The technical contribution of the paper is limited,'' and ``Besides, adding the binary feature in the embedding is not necessary; the LSTM model could learn such sequential correlation.'' The first two examples refer to broad aspects without providing enough detail for the authors to make improvements. In the third example, although the criticism initially seems valid, it is not supported by evidence. Therefore, we concluded that criticisms lacking helpful suggestions for future improvement or not supported by reasons should be considered non-constructive.

\paragraph{Remark.} 
We acknowledge that due to the lack of formal definition of toxicity and few literature are available, our toxicity definition in \S\ref{sec:guideline} can raise different opinion among researchers.  
However, we believe that if a review sentence falls into any categories in our toxicity guideline, it should attract extra attention and the reviewer who writes the sentence should be extra cautious. Furthermore, due to the subject nature of toxicity, the annotations can suffer from inconsistency, therefore, we have designed two stages of annotations and only include those sentences all reviewers are agreed in our final testing set.

\section{Benchmarks}
\label{sec:methods}

\paragraph{Task Formulation.}
Based on our created dataset, we formulate a binary classification task for toxic detection in the peer-review. The goal is to classify a sentence as either non-toxic or toxic. Any sentence that belongs to one of the four categories (\S\ref{sec:guideline}) will be classified as toxic. 
Note that, while we have the four sub-categories of toxicity, we decided to formulate the toxicity classification task as a binary classification for simplicity and more fine-grained classification will be an future work.

\subsection{Toxic Detection Models}

To understand whether existing automatic toxicity classifiers could be used to detect toxicity in peer review, we evaluate text in our dataset using the Perspective API\footnote{\url{https://perspectiveapi.com/}}. The Perspective API has been widely used in previous studies on automatic toxicity detection \cite{deshpande2023toxicity,faal2023reward,wen2023unveiling,howard2024uncovering} and outputs scores in the range of $(0,1)$ which quantify the likelihood of text containing various attributes related to toxicity. 
The model output 8 aspects: ``identity attack'', ``severe toxicity'',``profanity'', ``sexually explicit'',  ``toxcity'', ``threat'', ``insult'', and ``flirtation''.
We specifically use the Toxicity attribute returned by the Perspective API.

\subsection{Sentiment Analysis Models} 

As toxic peer reviews are also negative in nature, our task can be viewed as related to the problem of sentiment analysis~\cite{socher2013recursive}.
Consequently, we use a pre-trained sentiment classifier\footnote{\url{https://huggingface.co/lxyuan/distilbert-base-multilingual-cased-sentiments-student}} as another baseline for automatic detection of toxic reviews. The classifier returns separate scores in the range of $(0,1)$ quantifying the probability of a text sequence containing a positive, neutral, or negative sentiment. 
We use two way to decide if a sentence is toxic or not. 
SA-1: a sentence is toxic if the negative score is higher than the positive score, otherwise the sentence is non-toxic. 
SA-2:  a sentence is toxic if the negative score is higher than both positive and neutral score.

\subsection{Large Language Models} 
Large language models (LLMs) have shown significant promise in various natural language processing tasks. 
Therefore, we also extensively evaluate the state-of-the-art LLMs' performance on detecting toxic sentences in peer-review. We evaluate both open-source and closed-source models, covering a range of model sizes. For the open-source models, we evaluate Gemma-7B~\cite{team2024gemma}, Qwen-7B~\cite{bai2023qwen}, Mistral-7B~\cite{jiang2023mistral}, LLaMA3-8B, and LLaMA3-70B~\cite{touvron2023llama}, utilizing the model checkpoints available on the Hugging Face platform. For the closed-source models, we assess both GPT-3.5 and GPT-4.

\section{Experiments and Results}

\begin{table}[t]
\centering
 \resizebox{0.98\linewidth}{!}{
\begin{tabular}{c|c|c|c|c|c}
    \toprule
    \multirow{2}{*}{\textbf{Models}}& \multicolumn{5}{c}{\textbf{Performance}}\\
     \cmidrule(lr){2-6}
     & {\textbf{Precision}} &  {\textbf{Recall}}  & {\textbf{F1}} &  \textbf{Accuracy}  & \textbf{Cohen's Kapa}\\ 
     \toprule
    SA-1 & 55.89 & 47.28 & 42.01 & 47.28 & 0.04\\
    SA-2 & 57.76 & 55.59 & 55.80 & 55.59 & 0.12\\
    \midrule
    Gemma-7B & 53.32 & 52.40 & 52.78 & 52.40 & 0.04  \\
    Qwen-7B & 59.58 & 60.70 & 59.17 & 60.70 & 0.16 \\ 
    Mistral 7B & 53.83 & 50.48 & 50.66 & 50.48 & 0.05\\ 
     LLaMA-3 8B & 52.38 & 50.16 & 50.35 & 50.16 & 0.02 \\
     LLaMA-3 70B  & 51.95 & 49.52 & 50.00 & 49.52 & 0.01\\
     \midrule
     ChatGPT 3.5 & 75.95 & 43.45 & 28.09 & 43.45 & 0.02 \\
     ChatGPT 4 & 67.71 & 46.33 & 35.06 & 46.33 & 0.06\\
    \bottomrule
    \end{tabular}
    }
    \vspace{3mm}
\caption{Performance of Different Models on Toxic Peer-Review Detection Task with Simple Prompt.
}
\label{tab:detector_result}
\end{table}



\subsection{Toxicity Detection Performance and Analysis} 
\label{sec:experiment_performance}

\paragraph{Evaluation Metric.} Because of the in-balanced labels in the testing set, we report Precision, Recall, F1, Accuracy scores. Furthermore, we report Cohen's Kappa between the human label and each model performance.

\paragraph{Toxicity Detection Model.}
The toxicity detection models have predict very low probability of being toxic for all the sentences: the mean value cross the entire testing set is 0.03 and the max value is only 0.32. These probability is much lower than a threshold (e.g. 0.4) used in previous work. Meaning that the toxic semantic meaning in the peer-review context is largely different from the general domain. It is because usually a toxic peer-review does not contain overtly hostile or abusive language, which is often a defining characteristic of toxic reviews in general domain. Therefore, while it might be seen as problematic or unprofessional, it is not detected as ``toxic'' by a general toxic detection model. This emphasize the need for a new dataset and a specific model or model design for toxic peer-review detection task. 

\paragraph{Sentiment Analysis Model.}
We present the results of the two ways of toxicity prediction by a sentiment analysis model in Table~\ref{tab:detector_result}. 
The result show that the second way (i.e. the negative score is higher than both neural and positive) achieve higher alignment with human judgement. While the result does not reach a substantial alignment with human, it is even better than most of the open source LLMs which are at much larger parameters. 
This shows that negative sentiment is align with toxicity to some degree. 

\paragraph{Simple Prompt with LLMs.}
To prompt the LLMs, we first use a very simple and straightforward prompt as shown in Figure~\ref{fig:simple_prompt}. 
The results in Table~\ref{tab:detector_result} shows that Qwen-7B model achieve the best alignment with human compared to other open-source LLMs and even the close-source GPT-3.5 and GPT-4. 

\begin{figure}[h]
\lstset{frameround=fttt}
\begin{lstlisting}[frame=trBL,linewidth=1.01\columnwidth,breaklines=true,breakautoindent=false,breakindent=0pt,numbers=none]
Given a sentence from a peer-review, please judge if this sentence is toxic or not toxic.
Sentence: {sentence}
If it is toxic, generate label 1, otherwise, generate label 0. 1 or 0? Answer:
\end{lstlisting}
    \caption{Simple Prompt for LLMs to Conduct the Toxicity Detection Task.}
    \label{fig:simple_prompt}
\end{figure}

\begin{figure}
    \centering
    \includegraphics[width=0.95\linewidth]{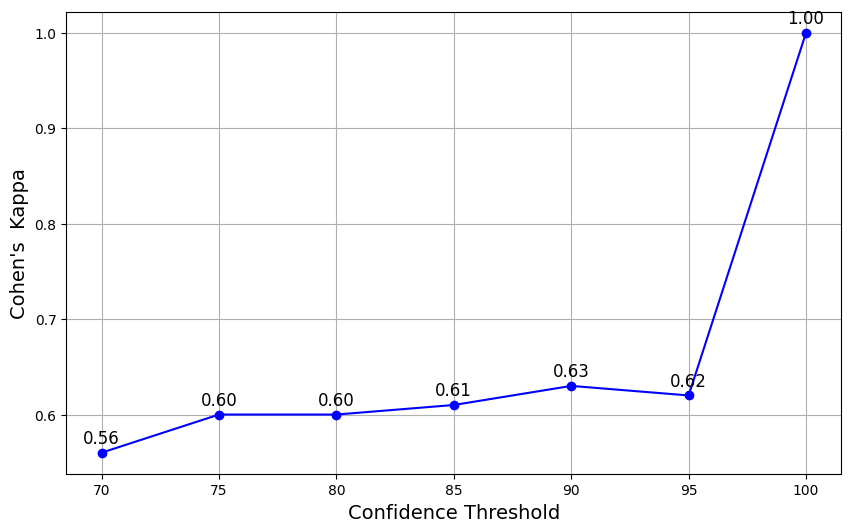}
    \caption{The Cohen's Kappa of GPT-4 Regarding to the Confidence Threshold.}
    \label{fig:confidence}
\end{figure}

\begin{table}[t]
\centering
 \resizebox{0.98\linewidth}{!}{
\begin{tabular}{c|c|c|c|c|c}
    \toprule
    \multirow{2}{*}{\textbf{Models}}& \multicolumn{5}{c}{\textbf{Performance}}\\
     \cmidrule(lr){2-6}
     & {\textbf{Precision}} &  {\textbf{Recall}}  & {\textbf{F1}} &  \textbf{Accuracy}  & \textbf{Cohen's Kapa}\\ 
     \toprule
     ChatGPT 3.5 & 68.19 & 62.62 & 62.14 & 62.62 & 0.28 \\
     ChatGPT 4 & 78.53 & 78.27 & 78.35 & 78.27 & 0.56 \\ 
     \midrule
     ChatGPT 3.5 & 69.06 & 45.69 & 33.45 & 45.69 & 0.05\\
     ChatGPT 4 & 73.96 & 65.81 & 64.96 & 65.81 & 0.35 \\
     
    \bottomrule
    \end{tabular}
    }
    \vspace{3mm}
\caption{Performance of GPT Models on Toxic Peer-Review Detection Task with Detailed Instruction Prompt (the upper block) and Toxicity Summary Prompt (the bottom block).}
\label{tab:prompt_detector_result}
\end{table}
\paragraph{Detailed Instruction with LLMs.} 
We provide more concrete definition of toxic peer-review sentence definition and extend the prompt with each subcategories definition. However, when given such detailed prompt to the open-source models, all models do not generate an intended answers (e.g. 0/1 or toxic/non toxic). Therefore, we do not present the performance for the open-source model with the detailed instruction. On the other hand, the close-source models, GPT-3.5 and GPT-4 can follow the instruction and achieve much better alignment compared to the previous simple prompt as shown in the first block performance in Table~\ref{tab:prompt_detector_result}. Encouragingly, the GPT-4 achieve 0.56 Cohen's Kappa score with human. 
Meanwhile, we also prompt the model to generate a confidence score of its answer. 
The min/max/mean values of the confidence is 70\%, 89\%, and 100\%, this shows that the model is quite confidence with its answer in most of the cases. 
We use the confidence to select the sentences to further compute the Cohen's Kappa Scores. 
As shown in Figure~\ref{fig:confidence}, choosing a higher confidence yield higher alignment and when the confidence is 100\%, the model reach a perfect alignment with the human judgement. 
However,  a higher threshold also means less sentences are being judged, the number of sentences being selected for the threshold shown in the figure are: 313, 300, 300, 258, 248, 53, 3. 

\paragraph{Toxicity Definition Instruction with LLMs.} 
The last prompt that we experiment is the a summary of the toxicity as shown Figure~\ref{fig:summary_prompt}. We compare this  result (bottom block) with the detailed instruction prompt result in Table~\ref{tab:prompt_detector_result}.
Both model performance decreases significantly, this showcase the importance of the detailed instruction for detecting the toxicity in the peer-review context. 

Meanwhile, we also prompt the model to generate a confidence score of its answer. 
The min/max/mean values of the confidence is 70\%, 89\%, 100\%, this shows that the model is quite confidence with its answer in most of the cases. 
We choose a high confidence threshold 90\% to select the cases to further compute the Cohen's Kappa Scores,

\begin{figure}[h]
\lstset{frameround=fttt}
\begin{lstlisting}[frame=trBL,linewidth=1.01\columnwidth,breaklines=true,breakautoindent=false,breakindent=0pt,numbers=none]
Peer review is vital to the scientific process, but feedback can sometimes have a negative emotional impact on paper authors. To be specific, sentences which feature emotive, rhetorical, narrativizing, universalizing, and/or subjective language, comments which lack substance, helpfulness, or specific, actionable guidance for improvement, personal attacks or author-focused critiques, and sentences which fixate excessively on the negative aspects of a paper are of particular concern.

\end{lstlisting}
    \caption{Toxicity Definition Summary}
    \label{fig:summary_prompt}
\end{figure}

\begin{table*}[t]
    \centering 
    \setlength{\tabcolsep}{4pt}
    \begin{tabular}{@{}p{0.4\linewidth}p{0.4\linewidth} p{0.2\linewidth} }
        \toprule
          \textbf{Original Sentence} & \textbf{GPT-3.5 Revision} & \textbf{Human Preference}\\
        \toprule
        I find this assumption absurd. & I respectfully disagree with this assumption. & Revision.\\
        \midrule 
        Overall the novelty of the approach and the proposed problem is incremental. & The approach and proposed problem show incremental novelty, which is a positive aspect of the study. & Original, the model revision change the meaning of the original sentences.\\
        \midrule 
        Notation and definitions used in (9) are wrong and confusing. & The notation and definitions used in (9) could benefit from clarification and correction to avoid confusion. & Revision \\
        \midrule
        There are so many errors like this and the paper needs serious rewriting. & There are several errors throughout the paper, and it would benefit from a thorough revision. & Revision\\
    \bottomrule
    \end{tabular}
    \caption{Examples of Detoxified Sentence from GPT.}
    \label{tab:detoxified_sentence}
\end{table*}

\begin{table*}[t]
    \centering 
    \setlength{\tabcolsep}{4pt}
    \begin{tabular}{@{}p{0.35\linewidth}p{0.3\linewidth}p{0.35\linewidth} }
        \toprule
         \textbf{Sentence} & \textbf{Annotator 1 Comment} &\textbf{Annotator 2 Comment} \\
        \toprule
        ` In fact, Figure 3 says it all,' - in fact, figure 3 does not say it all because it's a particular example projected into 2d without very much explanation. & Non Toxic & This seems like a sarcastic comment - I thought that might qualify as emotive comment. \\
        \hline
        While the selected baseline is important, it is not sufficient. & Non Toxic & I thought this was non-constructive because there's no elaboration on why it is insufficient \\
        \hline
        Why do you ignore your original problem and methods, and consider this problem?  & The way how the question being ask is quite emotive, it can be rephrased better & Non Toxic \\
        \hline
        However, in its current state I don't believe this is appropriate. & `I don't believe' sounds emotive & Non Toxic \\
    \bottomrule
    \end{tabular}
    \caption{Examples of Sentences Human Disagree With.}
    \label{tab:disagreement_sentence}
\end{table*}

\subsection{Toxicity Revision}

We also investigate if LLMs are capable of rewriting toxic sentences while maintaining the original critiques, an important aspect of ensuring constructiveness in the scientific review environment. We sample 10 sentences from each toxicity subcategory (with exception that personal attack type only has 5 sentences in the entire testing set), and prompt (Figure~\ref{fig:revision_prompt}) GPT-3.5 to revise them. 
Then we evaluate the revisions and judge whether the re-written sentences are less toxic compared to the original ones. We find that the revisions are favorable 80\% of the time (28/35 revisions) suggesting that the model is generally competent. we have provided more examples in Table~\ref{tab:detoxified_sentence}. 
For most of the ``lack of constructive feedback'', the model simply rephrase the sentence such as revise the original sentence ``Major baselines are missing.'' to ``The paper would benefit from including major baselines for a more comprehensive analysis.''. 
While the constructiveness does not improve, the annotator still reports that the revision is more polite and thus more preferable. On the other hand, it is almost not possible to make a comment more constructive without reading the paper, therefore, we suggest that for this type of toxicity, rather than ask the model to rewrite the sentence, the model should remind the reviewer to give more constructive feedback. 
\begin{figure}[h]
\lstset{frameround=fttt}
\begin{lstlisting}[frame=trBL,linewidth=1.01\columnwidth,breaklines=true,breakautoindent=false,breakindent=0pt,numbers=none]
This text is from a scientific paper review:
{sentence}
Revise this sentence such that it maintains the original critique but delivers it in a more friendly, professional and encouraging manner. Make minimal changes to the original text.
Your revision: 
\end{lstlisting}
    \caption{Revison Prompt.}
    \label{fig:revision_prompt}
\end{figure}

\subsection{Human Disagreement}
During our annotations, although the annotators have conducted detailed discussion, there are still some cases that they did not reach agreement. We exclude those data point in our final testing set, however, investigating these examples further can potentially be beneficial to improve our toxicity guideline. 
In Table~\ref{tab:disagreement_sentence}, we present such cases and the different comments from the annotators.

\section{Conclusion and Future Work}
Our work is the first to explore toxicity in peer review. Lacking a comprehensive guideline, we first defined toxicity and identified four key toxic categories. We developed a two-stage annotation process, ensuring our toxicity detection annotations are reliable and our final testing dataset includes only sentences with unanimous agreement among annotators. 
We then benchmarked various models, including a general toxic detection, a sentiment analysis model, and both open-source and closed-source large language models. Our results suggest that Open-source models struggled to align with human judgments, highlighting the challenge of detecting toxicity in peer reviews. Conversely, closed-source models like GPT-3.5 and GPT-4 showed much better alignment, suggesting their potential with careful use. Future work could involve using these models to generate synthetic data for fine-tuning open-source models.



\bibliography{custom}

\clearpage

\appendix


\section{Toxic Definition}

\begin{figure}[h]
    \centering
    \includegraphics[width=0.95\linewidth]{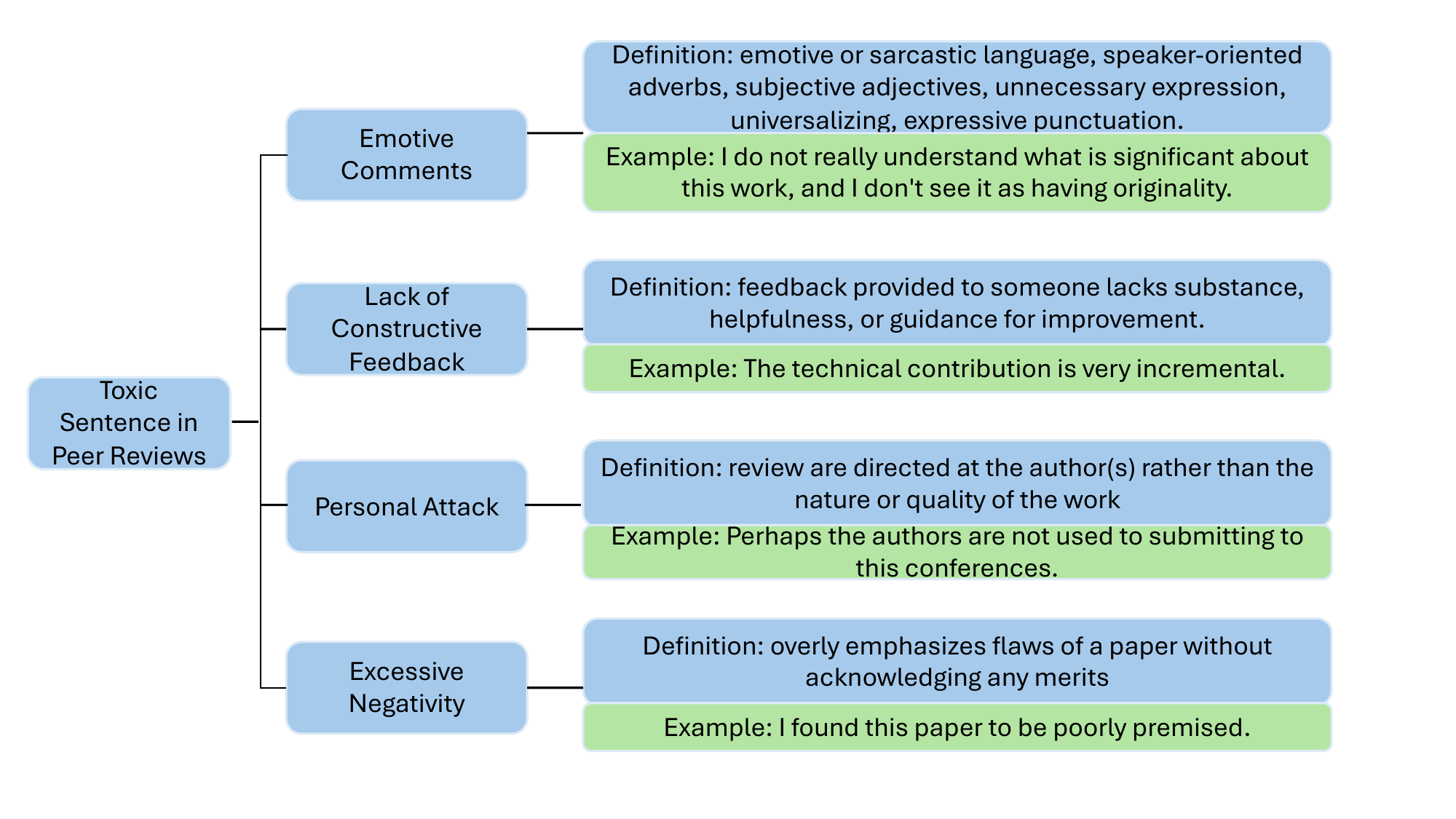}
    \caption{The Definition and Examples of Toxic Peer Review. }
    \label{fig:toxic_definition}
\end{figure}

In Figure~\ref{fig:toxic_definition}, we show the four sub-categories of our toxic definition and examples. 

\section{Conferences of the Reviews}

Our testing datasets are collected from 6 conferences or workshops and each one has the peer reviews public available on OpenReview Platform. 
In Table~\ref{tab:conference}, We list the conferences from which our testing set reviews were sourced.

\begin{table}[h]
\centering
 \resizebox{0.95\linewidth}{!}{
\begin{tabular}{c|c|c|c|c|c}
    \toprule
     ICLR & MIDL & ICAPS & Neurips & KDD & GI\\
     \midrule
     50 & 9 & 5 & 5 & 3 & 2 \\
    \bottomrule
    \end{tabular}
    }
    \caption{Toxic Review Detection Dataset. The statisic is in sentence level. GI for Graphics Interface.}
\label{tab:conference}
\end{table}

\section{General Toxicity Detection Model}

\begin{figure}
    \centering
    \includegraphics[width=0.95\linewidth]{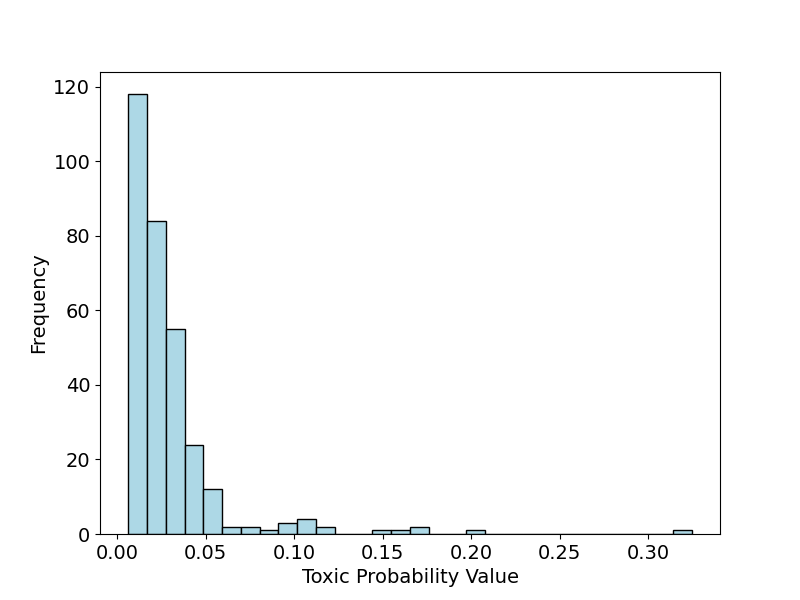}
    \caption{The Distribution of the Predicted Toxicity Probability Given by An Toxic Detection Model.}
    \label{fig:toxic_distribution}
\end{figure}

\begin{figure}
    \centering
    \includegraphics[width=0.95\linewidth]{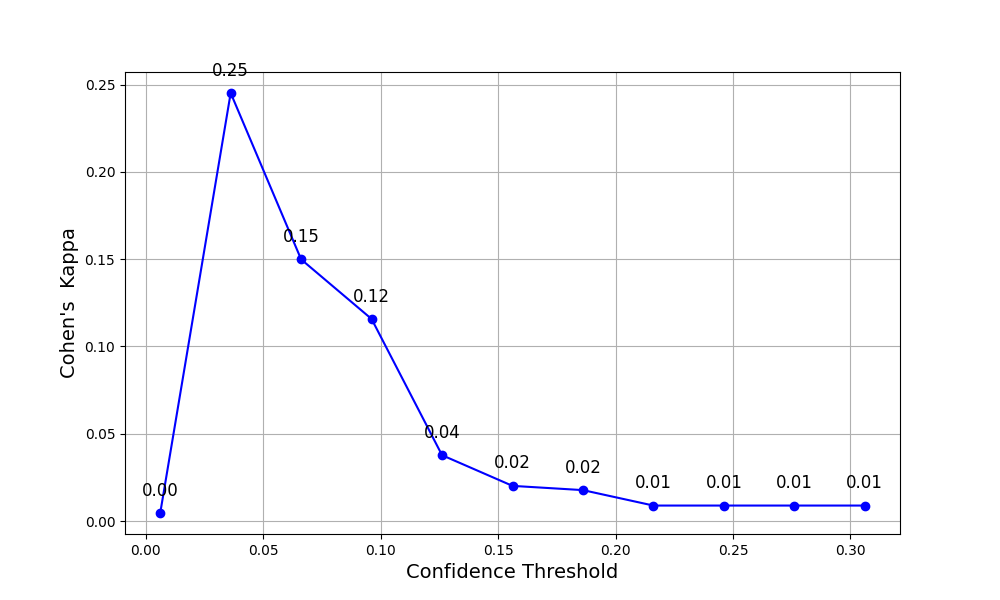}
    \caption{Enter Caption}
    \label{fig:threshold_toxic_model}
\end{figure}

As mentioned in \S\ref{sec:experiment_performance}, the toxicity detection model generates very low probabilities of toxicity for the sentences in our testing set. Figure~\ref{fig:toxic_distribution} illustrates the distribution of the predicted probabilities across the entire dataset, with the majority of values being less than 0.05. We further analyzed various threshold values to determine toxicity: if the predicted probability exceeds a given threshold, the sentence is classified as toxic; otherwise, it is classified as non-toxic. We began with a minimum threshold of 0.006, incrementing by 0.005 to establish subsequent thresholds. Figure~\ref{fig:threshold_toxic_model} depicts the relationship between the threshold and the Cohen's Kappa score. The maximum Cohen's Kappa score, approximately 0.25, falls within the range of [0.006, 0.06]. This suggests that potentially toxic peer-review sentences are generally assigned very low toxicity probabilities, indicating that the semantic interpretation of toxicity in peer review differs significantly from that in the general domain on which the toxicity model was trained.

\section{GPT-4 high Confidence Examples} 

As we mentioned in the main paper that the confidence is an good indicator for achieving good alignment with human judgement. Here we show the three examples where GPT-4 is 100\% confidence its answer and all of them are aligned with human judgement. 
Example 1 (Non-toxic): The random features are fixed once sampled from the base measure of the corresponding kernel.
Example 2 (Toxic): In short, the paper is not properly written nor well organized; is hard to read with vague contributions and vague positioning with respect to the state of the art.
Example 3 (Non-Toxic): The major experimental evaluations (Fig. 2 and Fig. 3) are based on the $m^2$ coverage after k steps and the plots are cut at 1000 steps.

\end{document}